\documentclass[preprint,12pt]{elsarticle}

\date{}

\usepackage{amssymb}

\usepackage{xcolor}
\usepackage{url}
\usepackage{enumitem}

\usepackage{booktabs}
\usepackage{tabularx}
\usepackage{multirow}
 
\usepackage[ruled,linesnumbered,vlined]{algorithm2e}

\usepackage{subcaption}
\usepackage{fancyhdr}

\newdefinition{definition}{Definition}

\let\oldmaketitle\maketitle
 
\renewcommand{\maketitle}{%
  \oldmaketitle
  \thispagestyle{fancy}
}

\journal{Robotics and Autonomous Systems}

\begin{document}

\begin{frontmatter}



\title{Receding Horizon Task and Motion Planning in Changing Environments}


\author[1,2]{Nicola Castaman\corref{cor}}
\ead{nicola.castaman@it-robotics.it}

\author[2]{Enrico Pagello}

\author[1]{Emanuele Menegatti}

\author[1]{Alberto Pretto}

\cortext[cor]{Corresponding author}

\address[1]{Department of Information Engineering, University of Padua, Padua, Italy}
\address[2]{IT+Robotics Srl, Vicenza, Italy}

\begin{abstract}
Complex manipulation tasks require careful integration of symbolic reasoning and motion planning. This problem, commonly referred to as Task and Motion Planning (TAMP), is even more challenging if the workspace is non-static, e.g. due to human interventions and perceived with noisy non-ideal sensors. This work proposes an online approximated TAMP method that combines a geometric reasoning module and a motion planner with a standard task planner in a receding horizon fashion. Our approach iteratively solves a reduced planning problem over a receding window of a limited number of future actions during the implementation of the actions. Thus, only the first action of the horizon is actually scheduled at each iteration, then the window is moved forward, and the problem is solved again. This procedure allows to naturally take into account potential changes in the scene while ensuring good runtime performance. 
We validate our approach within extensive experiments in a simulated environment. We  showed that our approach is able to deal with unexpected changes in the environment while ensuring comparable performance with respect to other recent TAMP approaches in solving traditional static benchmarks.
We release with this paper the open-source implementation of our method.
\end{abstract}


\end{frontmatter}


\section{Introduction}\label{sec:introduction}

A robot manipulator that manipulates objects in cluttered scenarios involves the solution of two fundamental sub-problems: determine the type and the order of the actions to be taken, and determine how to accomplish each action. The former problem is typically solved by high-level symbolic task planners, while the latter is addressed using a low-level geometric motion planner.
Many researchers are looking at these two problems as an intermingled problem to be solved simultaneously to discover feasible solutions efficiently. Thus, the two planners should not be used separately  and sequentially, but they should communicate to take into account both logic and geometric constraints. For example, task and Motion Planning (TAMP) problems are typically solved by calling the motion planner after the task planner to evaluate the feasibility of each action\cite{srivastava2014combined,dantam2016incremental,lagriffoul2012constraint} and, in case of infeasibility, sharing failure causes, or by closely intertwining task planning and motion planning to compute a full path plan~\cite{cambon2009hybrid,akbari2019combined,garrett2015ffrob}.

\begin{figure}[t]
    \centering
   \includegraphics[width=\columnwidth]{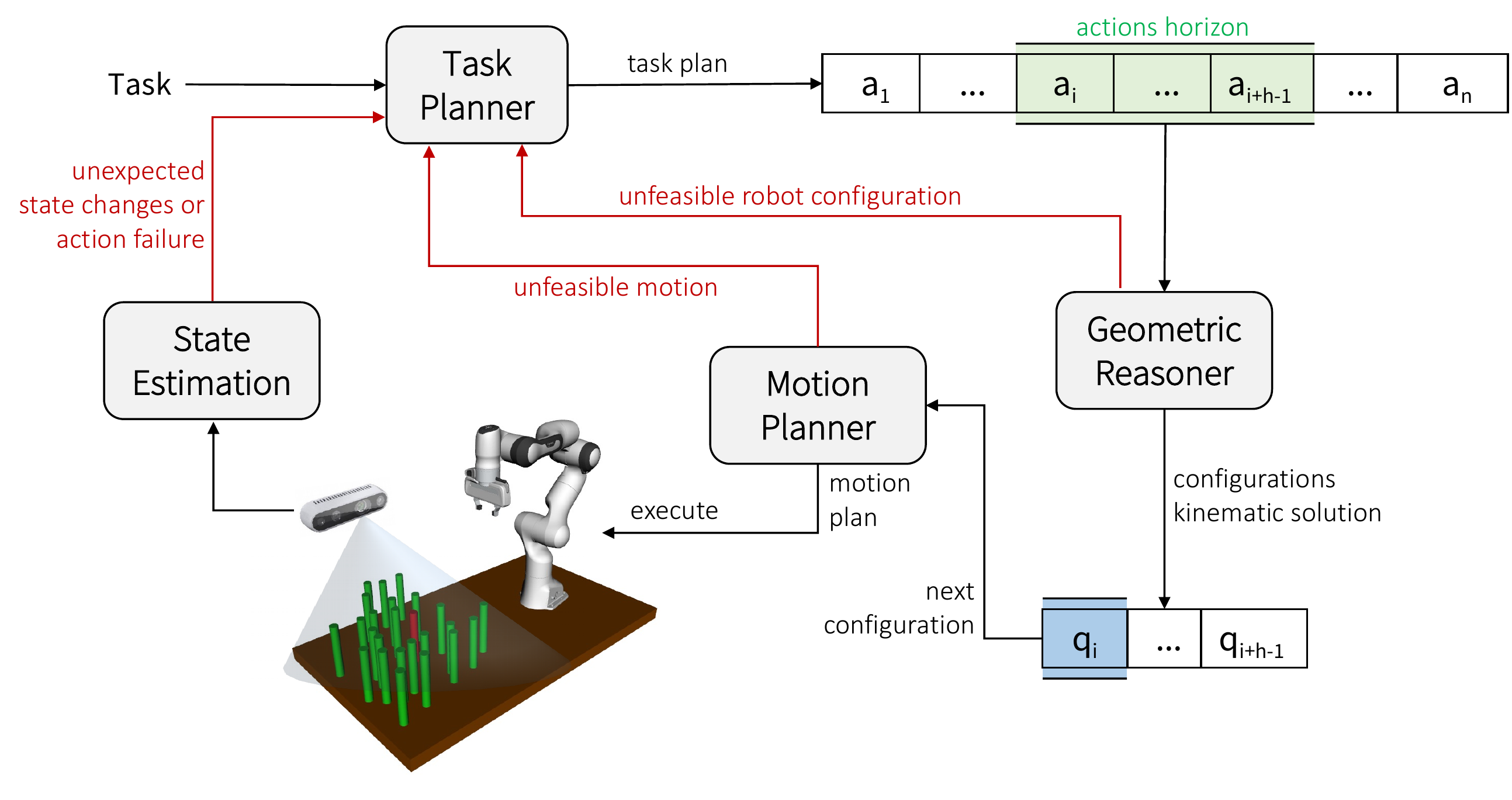}
    \caption{An overview of the proposed Receding Horizon TAMP method. Given a target task, a geometric reasoning module iteratively evaluates the feasibility of a sequence of $h$ actions (the \emph{actions horizon}) considered over the full set of $n$ actions provided by the task planner. Hence, the motion planner guides the robot toward the first available configuration $q_i$, and the process restarts by moving forward of one step the actions window of size $h$. If the plan is not feasible, the action fails, or the state of the system changes unexpectedly, the task planner should be executed again.}
    \label{fig:method_overview}
\end{figure}

Most current TAMP approaches assume to work in a static environment and to exploit ideal, noise-free sensors and actuators: the full plan is computed once at the beginning of the process and then performed assuming perfect actions. However, the environment in which the robot operates may undergo some changes during the action. In a cooperative robotic cell, for example, an operator collaborating with the robot for the same task could modify the environment (for example, by changing the position of some objects), invalidating in this way the plan computed in advance. Moreover, the perception of the environment could be incorrect due to sensors' noise and occlusions, while some actions may not be successful,  e.g. grasp failures. This work aims to relax the assumptions of static environment, ideal sensors, and ideal actuators, introducing an approximated TAMP approach that efficiently re-computes a sub-plan after performing each action. Thus, exploiting at each iteration updated sensory readings, it implicitly takes into account unexpected changes in the estimated state due to previous sensors and actuators failures or random environment re-configurations. Our approach, which we call \emph{Receding Horizon Task and Motion Planning} (RH-TAMP), takes inspiration from the Model Predictive Control (MPC) method \cite{GrnePannek2013}, a family of algorithms used to solve online optimal control problems over a receding horizon. In its original formulation, MPC uses a model of the process to be controlled to predict future outputs inside a finite window of future states (i.e., the \emph{horizon}). The differences between the predicted outputs and the desired references are minimized over this horizon \cite{Bemporad1999}, providing a sequence of optimal commands $u(t),\dots,u(t+h)$. Following a receding horizon strategy, at time $t$, only the first input $u(t)$ of the optimal command sequence is applied; in the next step, new measurements are collected, and the control problem is solved again, moving forward the finite horizon of one step. If the horizon is not too long, MPC is computationally efficient, and it can be exploited for real-time applications such as robot navigation and obstacle avoidance.\\

In RH-TAMP, we focus on effective and optimized integration of existing task planners and motion planners, as done  in other recent approaches \cite{srivastava2014combined,dantam2016incremental,akbari2019combined}. 
Unlike  these  methods, in RH-TAMP, we define an \emph{actions horizon} as a sub-sequence of the full actions sequence computed once by a task planner (Fig.~\ref{fig:method_overview}). Thus, the actions horizon starts from the first action not yet executed. Furthermore, our approach exploits a geometric reasoning module at each iteration to evaluate the feasibility of each action $a_i$ included in the horizon in terms of reachability and to compute the related robot configurations $q_i$. 
An action is classified as feasible by the geometric reasoner if there are  valid and collision-free robot configurations $q_i$ that enables reaching the target of the evaluated action.
At this point, similarly to MPCs, only the first action of the horizon is actually executed (e.g., action $a_i$ in Fig.~\ref{fig:method_overview}), exploiting a motion planner to guide the robot toward the related kinematic configuration ($q_i$ in Fig.~\ref{fig:method_overview}).
An action is classified as feasible by the motion planner if a collision-free path exists that the robot can follow to complete the corresponding action.
If either the geometric reasoning module or the motion planner module can't provide any solution, the task planner should be called again, and the process restarts from the new configuration (red solid arrows in Fig.~\ref{fig:method_overview}).
If we assume that the robot is coupled with an exteroceptive sensor like a camera or a depth sensor, by using a state estimation module, it can perceive in real-time the current scene configuration (e.g., the actual position of the objects in the workspace). If such configuration differs from the predicted output, i.e., the expected scene configuration after performing the actions, it means either the last action has failed or external agents have modified the scene. Also, in this case, the task planner should be called again. Otherwise, the process continues by moving forward the actions window, exploiting all the previous constraints.\\
As introduced, the proposed method provides an approximated solution to the TAMP problem. Such a solution could be sub-optimal, or in some cases, the algorithm could discover during the execution that an action of the original task plan is unfeasible and needs backtracking to reach the goal. Anyhow, for each failure, the algorithm always takes into account the cause, adding it as a new constraint to a list of \emph{logical predicates} used by the task planner. This approach makes it possible to avoid deadlock conditions deriving from re-planning a plan equivalent to a previous one.\\

We validated our method with extensive simulated experiments over three different TAMP benchmarks, including the \textit{Sort Clutter} problem involving a mobile manipulator, to demonstrate our approach's robustness to robot's localization errors.
Our results suggest that RH-TAMP is able to effectively solve non-static TAMP problems regardless of their cardinality while ensuring comparable performance with respect to other recent TAMP approaches in solving traditional, static problems. An open-source implementation of our system is made publicly available with this paper at\\
\hspace{1cm}\url{https://github.com/nicolacastaman/rh-tamp}.

\section{Related Work}

\subsection{Task Planning}
Task planning has been widely studied from early works on STRIPS~\cite{fikes1971strips}. Task planning approaches focus on efficiently searching the state space and are commonly based on Heuristic Search~\cite{hoffmann2003metric,helmert2006fast} and Constraint-Based methods~\cite{kautz1999unifying,rintanen2012planning,rintanen2014madagascar}.
Heuristic-Search methods, such as Fast-Forward (FF)~\cite{hoffmann2003metric}, use heuristics to reduce the number of states to be expanded. Constraint-Based methods propagate constraints to avoid searching the entire state space.

Task domains are usually represented with Planning Domain Definition Language (PDDL)~\cite{mcdermott1998pddl,fox2003pddl2,edelkamp2004pddl2}. PDDL tries to standardize the setup of AI planning problems.

In our work, we define problems in PDDL and exploit FF for task planning resolution, but other standard task planners could be used instead with only minor modifications.

\subsection{Motion Planning}
Motion planning algorithms for robots with a large number of degrees of freedom (DoF) are mainly based on Sampling~\cite{kavraki1996probabilistic,lavalle1998rapidly,kuffner2000rrt} or Optimization~\cite{schulman2013finding,zucker2013chomp,kalakrishnan2011stomp} methods.
Sampling-Based approaches efficiently handle high DoF manipulators. A typical approach is Probabilistic Roadmap Methods (PRM)~\cite{kavraki1996probabilistic} which uses a probabilistic approach to generate a roadmap that covers the free configuration space.
An alternative approach is Rapidly Exploring Random Tree (RRT)~\cite{lavalle1998rapidly}. RRT explores the configuration space by expanding several branches of a tree. RRT-Connect~\cite{kuffner2000rrt} is a variant of RRT in which the root of trees are located at the start and goal configuration and try to meet each other.

Optimization-Based approaches~\cite{zucker2013chomp,kalakrishnan2011stomp,schulman2013finding} can quickly solve some motion planning problems, but they require a good initial guess; these planners may also encounter some difficulties in case of narrow passages and small obstacles. But, on the other hand, they can be used profitably to smooth and shorten the trajectories generated by other methods.

Our work uses RRT-Connect, but other standard sampling-based planners could be used instead with only minor modifications.

\subsection{Task and Motion Planning}

Combination of task and motion planning needs to search valid actions in symbolic space that are feasible in geometric space.

A common technique used to solve TAMP problems is to interleave the symbolic and geometric search processes by calling a motion planner at each step, to assign geometric parameters to the currently symbolic state before advancing to the next one. As an example, the aSyMov planner, presented in Cambon et al. ~\cite{cambon2004robot,cambon2009hybrid}, uses an FF-based task planner with lazily-expanded roadmaps. 
However, interleaving symbolic search with geometric search becomes problematic when a planned state is valid in symbolic space but geometrically infeasible.
To address this issue, Dornhege et al.~\cite{dornhege2012semantic} call the motion planner after each action, executing a feasibility check with the introduced \textit{semantic attachments}, which are external reasoners called when a state is evaluated. 
Garrett et al. with FFRob~\cite{garrett2015ffrob} introduce an FF-like heuristic that incorporates geometric information into FF-search.
Kaelblin and Lozano-Pérez~\cite{kaelbling2011hierarchical,kaelbling2013integrated} propose the Hierarchical Planning in the Now method (HPN), that interleaves planning with execution using a hierarchical approach so that actions are executed as soon as the search algorithm reaches them. This approach requires reversible actions when backtracking is necessary.
Similarly, de Silva et al.~\cite{desilva2013towards,desilva2013interface} exploits Hierarchical Task Networks (HTNs) to perform a symbolic search using hierarchically abstracted tasks. HTNs use shared literals to control backtracking between the task and motion layer.

The integration of symbolic search with geometric search can considerably reduce the symbolic space, but calling the motion planner after each symbolic search can take a long time if most states are geometrically feasible.
An alternative approach is to perform a geometric search only on full candidate symbolic plans.
Srivastava et al.~\cite{srivastava2014combined} interface a task planner with an optimization-based motion planner and use a heuristic to remove occluding objects. Lozano-Pérez and Kaelbling~\cite{lozano2014constraint} formulated the motion part as a constraint-based problem. Similarly, Dantam et al.~\cite{dantam2016incremental} use an incremental Satisfiable Modulo Theory (SMT) solver to incrementally generate symbolic plans and invoke a motion planner in between for validation.

The TAMP methods described above usually require a long processing time, from tens of seconds to minutes.
To address this issue, Wells~\emph{et al.}~\cite{wells2019learning} use a Support Vector Machine (SVM) to estimate the feasibility of actions to guide the symbolic search, only calling the motion planner on symbolic plans classified by the SVM as feasible.
In a similar way, Akbari et al.~\cite{akbari2019combined} introduce the concept of Geometric Reasoning used to verify the feasibility of actions\footnote{In this context the action is defined as feasible only by considering its final configuration and without considering the complete movement to perform that action.}, calling the motion planner only for feasible actions. 

Most of the introduced TAMP approaches take into account a static environment while leveraging an ideal, noise-free perception system and performing deterministic actions. Suárez-Hernández et al.~\cite{Suarez_IROS2018} attempted to overcome perception errors by incorporating a symbolic action in which the robot examines an object closely when the uncertainty of perception is high. 
However, this approach is still unable to handle environmental changes, such as objects being moved from their original position.
Migimatsu and Bohg \cite{Migimatsu_RAL2020} proposed to plan over relative object poses, so keeping a valid plan also if the object moves. Unlike these methods, in our approach we explicitly take into account the non-static nature of the environment and the non-deterministic nature of actions and perception by iteratively matching the desired and perceived outputs.

\subsection{Contributions}

Our contributions are the following: 
\begin{itemize}
    \item A novel online TAMP approach for changing environments based on an iterative, finite-horizon re-planning strategy;
    \item An extensive experimental evaluation in simulated environments;
    \item An open-source implementation of the proposed method developed within a custom-built general-purpose TAMP simulation framework.
\end{itemize}

\section{TAMP Background}
This section provides a basic theoretical formulation of the TAMP problem, defining the involved domains and the sub-problems addressed in the next section.

\begin{definition}[Task Domain]
A Task Domain is a tuple $\Sigma = (S, A, \gamma, s_0, S_G)$, where:
\begin{itemize}
    \item $S$ is a finite set of states
    \item $A$ is a finite set of actions
    \item $\gamma: S \times A \to S$ is a deterministic state-transition function that gives a new state when applicable. We represent it with $ \gamma(s_i, a) = s_{i+1}$, where $s_i,s_{i+1} \in S$ and $a \in A$.
    \item $s_0 \in S$ is the initial state
    \item $S_G \subseteq S$ is the finite set of accepted states (\textit{i.e.}, the task goal)
\end{itemize}
\end{definition}

\begin{definition}[Task Plan]
A Task Plan $\mathbf{A}$ is a sequence of actions $\mathbf{A} = \langle a_1, a_2, ..., a_{n} \rangle$ where each $a_i \in A$, $s_{i+1} = \gamma(s_i, a_{i+1})$, and $s_n \in S_G$.
\end{definition}

A robot manipulator can be modelled as a kinematic chain or kinematic tree of joints and links \cite{hartenberg_denavit_1964}. Its configuration can be represented by a vector of generalized joint coordinates $q \in \mathcal{C}$, the robot configuration space.

\begin{definition}[Motion Domain]
A Motion Domain is represented by the robot configuration space $\mathcal{C}$, i.e., the set of all possible configurations $q_i$ the robot may attain. The free configuration space, $\mathcal{C}_\mathrm{free} \subseteq \mathcal{C}$, is the space of all possible configurations that let the robot move from an initial configuration $q_I$ to a goal configuration $q_G$ while avoiding collisions with objects or self-collisions.
\end{definition}

\begin{definition}[Motion Plan]
A motion plan can be defined as either:
\begin{itemize}
\item A sequence of robot configurations $Q = \langle q_0, q_1, ..., q_m \rangle$, where $q_0 = q_I$ the initial configuration and $q_m = q_G$ a goal configuration, each $q_i \in \mathcal{C}_\mathrm{free}$, and the distance between subsequent configuration is small $\| q_{i+1} - q_i \| < \epsilon$.
\item A continuous trajectory $\tau : [0,1] \leftarrow \mathcal{C}_\mathrm{free}$ such that $\tau(0) = q_I$ and $\tau(1) = q_G$.
\end{itemize}
\end{definition}

A motion planning algorithm finds valid plans $\mathbf{Q}(q_I, q_G)$ from $q_I$ to $q_G$ over a configuration space $\mathcal{C}$.\\

\begin{definition}[Task and Motion Domain]
A Task and Motion Domain is a tuple $D = (\Sigma, \mathcal{C}, \phi, \xi, q_0)$, where,
\begin{itemize}
    \item $\Sigma$ is a task domain
    \item $\mathcal{C}$ is the robot configurations space
    \item $\phi$ is a function that maps states into robot configurations
    \item $\xi$ is a function that maps actions into motion plans
    \item $q_0 \in \mathcal{C}$ is the initial robot configuration
\end{itemize}\end{definition}

\begin{definition}[Task and Motion Plan]
A Task and Motion Plan $\mathbf{T}$ is a sequence of actions and motion plans pairs $\mathbf{T} = \langle (a_0, Q_0), (a_1, Q_1), ... (a_n, Q_n) \rangle$.
For each pair $(a_i, Q_i)$, $Q_i$ corresponds to a valid motion plan for the action $a_i$. And $first(Q_0)=q_0$, for $0<i<n$ $last(Q_i) = first(Q_{i+1})$, and $last(Q_n)=q_n$, with $q_n$ being final configuration.
\end{definition}

The Task and Motion Planning problem requires strong relationships between states, configurations, actions, and motion plans \cite{lagriffoul2018platform}. For example, states of the Task domain should involve feasible robot configurations; a Task action $a$ can be performed only if there exists a valid motion plan $Q \in \mathcal{C}_\mathrm{free}$ that enables the required state transition.

\section{RH-TAMP}

This work focuses on the Task and Motion Planning problem for robot manipulator workings in a non-deterministic, partially observable environment. Non-deterministic means that the state could change due to external, unexpected actions; in addition, the actions executed by the robot (for example, object grasping and placement operations) may fail. Partially observable in our settings means that the robot is able to perceive at least all the aspects of the environment relevant to accomplish the target task. Still, the perception process is performed using noisy and non-ideal sensors. To deal with the stochastic nature of the problem, we propose RH-TAMP, an iterative, approximated approach inspired by the Model Predictive Control theory (see also Sec.~\ref{sec:introduction}).\\

The entry point of our method (see Fig.~\ref{fig:method_overview}) is a task planner module that quickly determines the type and the order of the actions $\mathbf{A} = \langle a_1, a_1, ..., a_{n} \rangle$ to be taken to accomplish the target task.
Similar to other approaches (e.g.,~\cite{akbari2019combined}), we introduce a geometric reasoning module (see Sec.~\ref{sec:geom_reasoning}) that allows us to quickly evaluate the feasibility of $\mathbf{A}$ and compute the robot configurations $q_i$ corresponding to the states $s_{i} = \gamma(s_{i-1},a_{i})$, avoiding to waste time on planning motions for non-feasible trajectories. Actually, in basic TAMP approaches, most of the time is spent on planning motions of actions that are infeasible. In our RH-TAMP approach, at each iteration, the geometric reasoning module evaluates the feasibility of a sub-sequence of $h$ actions (the \emph{actions horizon}, see Fig.~\ref{fig:method_overview}), finding a set of valid kinematic solutions (if any) for each action.

The size of the horizon represents a trade-off between optimality and computational efficiency. For example, a window covering the whole plan $\mathbf{A}$ is useful for ensuring the geometric feasibility of the plan but at a higher computational cost. Conversely, to keep the system fast and responsive to changes in the environment, a small horizon $h$ (e.g., 2 or 3 actions) is preferable.
If the actions horizon is feasible, similarly to MPCs, only the first action of the horizon is executed ($a_i$ in Fig.~\ref{fig:method_overview}): to this end, a motion planner provides the sequence of motions that brings the robot to the target configuration ($q_i$ in Fig.~\ref{fig:method_overview}). 
After performing each action, a perception module connected to an exteroceptive sensor provides an estimate of the state of the system. If the perceived state (e.g., the positions of the objects) matches the desired state, it means that the action was successful and led to a correct change in the state. In this case, the action horizon is moved forward of one step, and a new iteration begins with the geometric reasoning evaluation of the new action entered into the horizon, i.e., the re-planning in the new horizon will take place by reusing most of the geometric reasoning carried out in the previous iteration. Otherwise, it means that:
\begin{itemize}
    \item The action has failed, or;
    \item Sensory information is noisy or misinterpreted, or;
    \item A change in the configuration of the environment has been unexpectedly carried out by an external agent.
\end{itemize}
In this case, the task plan is no longer valid, and the task planner should compute a new plan considering the new state configuration (red solid arrows in Fig.~\ref{fig:method_overview}). A task re-planning is also required in case of failure of either the geometric reasoning or the motion planner. These modules also report the cause of the failure, e.g., the colliding objects (red arrows in Fig.~\ref{fig:method_overview}). Possibly colliding objects and other causes of infeasibility are coded online as logical predicates and added to the task planning problem; the algorithm restarts right from the task planning. 
Note that at each iteration of the algorithm, the motion planner is called to find a valid trajectory to complete only the next action of the plan (e.g., move the gripper close to a specific object). In other words, the motion planner allows to reach the next robot configuration $q_i$ in Fig.~\ref{fig:method_overview}: this allows to save time in planning the motion in the full horizon.
The fact that for each iteration, only the first action of the horizon is actually executed has a twofold benefit: (i) it allows to react promptly in case of changes in the environment or action failures;
(ii) it avoids executing a long sequence of actions that turn out to be infeasible once considered subsequent actions following the current horizon. 

\subsection{RH-TAMP Algorithm}

\begin{algorithm}[!ht]
\KwIn{$D = (\Sigma, \mathcal{C}, \phi, \xi, q_0)$, $h$}

$s \leftarrow null$\; 
$s'\leftarrow$ \texttt{updateState}()\;
\While{$s' \notin S_G$}{\label{alg:outer_loop}
    \If{$s \neq s'$}{\label{alg:state_mismathc}
        $s \leftarrow s'$\; 
        $P \leftarrow \emptyset$\; 
        $f$ $\leftarrow$ false\;
    }
    \If{$\lnot f $}{\label{alg:task_replan}
        $\mathbf{A}$ $\leftarrow$ \texttt{taskPlan}($\Sigma$, $s$, $P$)\;\label{alg:tp}
        $f$ $\leftarrow$ true\;
        $i$ $\leftarrow$ 1\;
    }
    
    \ForEach{$a_j$ in $\mathbf{A}$ where $j \in (i, i + h - 1)$}{
      $(q_{j}, p_j) \leftarrow$ \texttt{geometricReasoning}($a_j$)\;\label{alg:gr}
      \If{$\lnot q_j$}{\label{alg:ac}
        $f$ $\leftarrow$ false\;
        $P$ $\leftarrow$ $P \cup p_j$\;
        break\;
      }
    }
    \If{f}
    {
      $(Q_i, p_i)$ $\leftarrow$ \texttt{motionPlan}($q_i$,$q_{i+1}$)\;\label{alg:mp}
      \eIf{$\lnot Q_i$}{\label{alg:ac2}
        $f$ $\leftarrow$ false\;
        $P$ $\leftarrow$ $P \cup p_i$\;
      }{
        \texttt{execute}($Q_i$)\;\label{alg:exec}
        $s  \leftarrow  \gamma(s, a_i)$\;\label{alg:gamma}
        $i \leftarrow i+1$\;
      }
    }
    $s'\leftarrow$ \texttt{updateState}()\;\label{alg:update_step}
}
\caption{RH-TAMP}
\label{alg:tmp}
\end{algorithm}

Algorithm~\ref{alg:tmp} describes the main steps of the RH-TAMP approach. The input is a Task and Motion tuple $D$ and the size $h$ of the actions horizon. The system state $s$ is initialized with an invalid state while the actual state $s'$ is observed by a perception system. An outer loop (block from line~\ref{alg:outer_loop}) continues until the actual state $s'$ is in the set of accepted states $S_G$. If $s'$ does not match the current, desired state $s$, that is the result of an action application (i.e., $s_i = \gamma(s_{i-1}, a_i)$), $s$ is reset to $s'$, the set of logical predicates $P$ is cleared, and a task re-planning is requested (\textit{if} statement block from line~\ref{alg:state_mismathc}).
If a task re-planning is required, the task planner is called, providing as inputs the new initial state $s$ and the accumulated set of predicates $P$ (\textit{if} statement block from line~\ref{alg:task_replan}). 
A geometric reasoning evaluation (Sec.~\ref{sec:geom_reasoning}) is performed for each action included in the currents actions horizon (line~\ref{alg:gr}); if some action is not feasible, the geometric evaluation is aborted, a new symbol is added to the list of predicates $P$, and a task re-planning is requested (\textit{if} statement block from line~\ref{alg:ac}). If the geometric reasoning is successful, the motion planner is called (line~\ref{alg:mp}). 
If the motion is not feasible, a new predicate $p_i$ that codes the reason of failure is added to the list of predicates $P$, and a task re-planning is requested (\textit{if} statement block from line~\ref{alg:ac2}). Otherwise, the motion plan is executed (line~\ref{alg:exec}), and the current, desired state is updated with the state-transition function $\gamma$ (line~\ref{alg:gamma}). Finally, at each iteration the actual state $s'$ is observed by a perception system (line~\ref{alg:update_step}).

\subsection{Geometric Reasoning}\label{sec:geom_reasoning}

The geometric reasoning module (line~\ref{alg:gr}) acts as a fast motion planner's approximator and has been implemented similarly as proposed in \cite{akbari2019combined}. It aims to geometrically synthesize and test the feasibility of a task plan from a geometric point of view. Specifically, (a) it maps an action $a_i$ (e.g., ``place the object $o$ onto table $t$'') into a geometric domain (``the target position of the object $o$ in table $t$ is ($x,y,z$), reached with the robot configuration $q_i$'') while (b) testing that such configuration is geometrically feasible (i.e., $q_i \in \mathcal{C}_\mathrm{free}$). The task (a) can be defined as \emph{Spatial Reasoning}, the task (b) as \emph{Reachability Reasoning} and they can be solved by calling an Inverse Kinematic solver and testing the possible collisions with any other object. If no valid configurations can be computed, the reasoner reports a failure and the cause that generated it (e.g., a colliding  object). In this case, we can avoid calling the motion planner. Conversely, $q_i$ represents a valid goal configuration; at this point, the motion planner can be called to test if and how the robot can move from its current configuration to $q_i$.

The geometric reasoning module internally synthesizes the final configuration of the robot and, possibly, the final position of the object the robot is manipulating. Doing so, it should consider the full 3D structure of the environment that could include movable and fixed objects to check if the tested configuration is collision-free.

\section{Implementation Details}
This section provides technical information about the RH-TAMP system, including the definition of the motion primitives used in the task planning and the implementation choices we made. Motion primitives are presented both for a manipulator and a mobile manipulator.

\subsection{Motion Primitives}\label{sec:primitives}

\subsubsection{Manipulation Primitives}

In our formulation, the actions used in the addressed manipulation problems are \textit{Pick}, \textit{Place}, \textit{Stack}, and \textit{Unstack}. \textit{Pick} and \textit{Place} actions are used to manipulate objects over a surface; multiple objects can be placed on a surface. \textit{Stack} and \textit{Unstack} actions are used to manipulate objects over other objects or a precise location over a surface; only one object can be stacked on top of another object.

The \textit{pick(a,b)} action allows the manipulator to pick up the movable object $a$ from a surface $b$. The preconditions are that the robot is currently holding no other objects, and no other objects obstruct $a$. The effect is that $a$ is no more placed on the surface $b$. The action is also responsible for negating $obstructs$ facts for the associated $a$ if any.
The symbolic definition is:
\begin{description}[labelindent=0.5cm]
\item [Parameters:] $a - object, movable$; $b - surface$
\item [Preconditions:] $\nexists o - object \mid inHand(o) \;\land\; \nexists o - object \mid obstructs(o,a)$
\item [Effects:] $inHand(a) \;\land\;\forall o - object \mid \neg obstructs(a,o) \;\land\;\neg on(a,b)$
\end{description}

The \textit{place(a,b)} action allows the manipulator to place the movable object $a$ over the surface $b$.
The preconditions are that $a$ must be in the end-effector and no other objects obstruct the surface $b$. The effects are that the end-effector is no more holding $a$ and $a$ is on $b$. 
The symbolic definition is:
\begin{description}[labelindent=0.5cm]
\item [Parameters:] $a - object, movable$; $b - surface$
\item [Preconditions:] $inHand(a) \;\land\; \nexists o - object \mid obstructs(o,b)$
\item [Effects:] $\neg inHand(a) \;\land\; on(a,b)$
\end{description}

The \textit{unstack(a,b)} action allows the manipulator to unstack the movable object $a$ from the object $b$. The preconditions are that the robot is currently holding no other objects, and no other objects obstruct $a$. The effect is that $a$ is no more placed on the object $b$. The action is also responsible to negate $obstructs$ and $leaveClear$ facts for the associated $a$, if any. The $leaveClear(a, b)$ fact avoids stacking an object over another object or surface $a$ in order to reach the object $b$.
The symbolic definition is:
\begin{description}[labelindent=0.5cm]
\item [Parameters:] $a - object, movable$; $b - object$
\item [Preconditions:] $\nexists o - object \mid inHand(o) \;\land\; \nexists o - object \mid obstructs(o,a)$
\item [Effects:] $inHand(a) \;\land\;\neg on(a,b) \;\land\; (\forall o - object \mid \neg obstructs(a,o) \;\land\;\neg leaveClear(o,a))$
\end{description}

The \textit{stack(a,b)} action allows the manipulator to stack a movable object $a$ over the object $b$.
The preconditions are that the robot is currently holding no other objects, no other objects obstruct $a$ or are stacked on $b$, and that $b$ has not to be left clear.
The effects are that the end-effector is no more holding $a$ and $a$ is stacked on $b$.
The symbolic definition is:
\begin{description}[labelindent=0.5cm]
\item [Parameters:] $a - object, movable$; $b - object$
\item [Preconditions:] $\nexists o - object \mid inHand(o) \;\land\; \nexists o - object \mid on(o,b) \;\land\; \nexists o - object, movable \mid leaveClear(b,o) \;\land\; \nexists o - object \mid obstructs(o,b)$
\item [Effects:] $\neg inHand(a) \;\land\; on(a,b)$
\end{description}

\subsubsection{Mobile Manipulation Primitives}

Similarly to the manipulation primitives, the actions used in the addressed mobile manipulation problems are \textit{Pick}, \textit{Place}, \textit{Load}, \textit{Unload}, and \textit{Navigate}. Again, \textit{Pick} and \textit{Place} actions are used to manipulate objects over a surface but require that the robot is near the surface. \textit{Load} and \textit{Unload} actions are used to put an object on the robot or remove it. In the end, \textit{Navigate} is used to move the robot between different surfaces.

 The \textit{pick(m, s)} action allows the mobile manipulator to pick up the movable object $m$ from a surface $s$, when near $w$. No object must be in the robot hand, and no object must obstruct $m$. As an effect, the robot holds $m$, and $m$ does not obstruct any object.
 The symbolic definition is:
    \begin{description}[labelindent=0.5cm]
    \item [Parameters:] $m - object, movable$; $s - surface$;
    \item [Preconditions:] $\nexists c - object \mid holding(c) \;\land\; on(m, s) \;\land\; \exists w - region \mid (in(s, w)\;\land\;near(w))\;\land\;   \nexists c - object \mid obstruct(c, m)\;\land\;   \nexists c - object \mid on(c, m)$
    \item [Effects:]  $holding(m) \;\land\; \neg on(m, s)\;\land\; \forall c - object \mid \neg obstruct(m, c) \;\land\; \forall w -region \mid \neg in(m,w)$
    \end{description}
 
The \textit{place(m, s)} action lets the mobile manipulator place $m$ from the robot hand to the surface $s$. 
The action is possible if no object obstructs the drop location $d$, and the robot is near the corresponding $w$. At the end, $m$ is on $d$:
    \begin{description}[labelindent=0.5cm]
    \item [Parameters:] $m - object, movable$; $d - surface, drop$;
    \item [Preconditions:] $holding(m) \;\land\; \exists w - region \mid (in(d, w) \;\land\; near(w)) \;\land\;
    \nexists c - object \mid obstruct(c, d)$
    \item [Effects:] $\nexists c - object \mid holding(c) \;\land\; on(m, d) \;\land\; \forall w -region \mid (in(d,w)\Rightarrow in(m,w))$
    \end{description}
    
The \textit{load(m)} action allows the robot to load \textit{m} into its tray. The action is feasible if the robot is holding \textit{m}. As effect, \textit{m} is in the tray:
    \begin{description}[labelindent=0.5cm]
    \item [Parameters:] $m - object, movable$;
    \item [Preconditions:] $holding(m)$
    \item [Effects:] $\nexists c - object \mid \; holding(c) \;\land\; loaded(m)$
    \end{description}

The \textit{unload(m)} action allows the robot to unload \textit{m}:
    \begin{description}[labelindent=0.5cm]
    \item [Parameters:] $m - object, movable$; 
    \item [Preconditions:] $\nexists c - object \mid holding(c) \;\land\; loaded(m)$
    \item [Effects:] $ holding(m) \;\land\; \neg loaded(m)$
    \end{description}

The \textit{navigate(x, y)} action lets the mobile manipulator move from a region $x$ to a region $y$. The robot should be near $x$ and its hand should be empty. It ends the motion near $y$:
    \begin{description}[labelindent=0.5cm]
    \item [Parameters:] $x - region$; $y - region$
    \item [Preconditions:] $\nexists c - object \mid \; holding(c) \;\land\; near(x)$
    \item [Effects:] $\neg near(x)\;\land\; near(y)$
    \end{description}

\subsection{Implementation}

The proposed method is developed within a custom-built general-purpose TAMP simulation framework.
Similarly to other state-of-the-art works~\cite{srivastava2014combined,garrett2015ffrob,dantam2016incremental}, our approach combines out-of-shelves task planners and motion planners.
Task planning is implemented exploiting the Fast-Forward (FF)~\cite{hoffmann2003metric} planner.
Some functionalities of the Geometric Reasoning module and the Motion Planning are implemented using MoveIt!~\footnote{\url{https://moveit.ros.org/}}, an open-source tool for robotics manipulation. MoveIt! under the hood integrates the Open Motion Planning Library (OMPL)~\cite{sucan2012the-open-motion-planning-library}, a motion planning library. OMPL provides the implementation of several sampling-based motion planning algorithms: in this work, the RRT-Connect motion planner is used.
MoveIt! also integrates the Kinematics and Dynamics Library (KDL) for kinematics and a collision detection modules to detect collisions between the robot and objects.
Finally, Djikstra’s algorithm and the Dynamic-Window Approach (DWA)~\cite{fox1997dynamic} are used for the navigation of the mobile manipulator.
The overall TAMP simulation framework is implemented on the Robotic Operating System (ROS)~\cite{quigley2009ros}.

Although we use a specific task planner and motion planner, the proposed framework is agnostic with respect to them. It is, in fact, possible to use other task planning algorithms (e.g. Fast Downward) by creating a wrapper or selecting any motion planning algorithm implemented in MoveIt!.

Geometric details of symbolic actions (i.e., object placements or robot grasp configurations) are not pre-computed. The geometric reasoner samples and assigns them on demand during the planning process.
Grasping points are sampled around the object to be manipulated or extracted from a pre-computed set. Potential grasping poses are validated by calling an Inverse Kinematic solver (IK) and checking whether the IK solution is collision-free or not.
The addressed problems restrict the robot to perform only side-grasps to manipulate the objects. Therefore in this implementation, grasping points are sampled only around the z-axis of the objects.
Similarly,  the positions of the objects are computed during the problem solution. 
For the object involved in the manipulation action, a set of place positions are sampled on the target surface; they are validated by calling a collision detection module checking if the sampled location is collision-free. In the case that a valid place position is found, the geometric reasoner verifies that a collision-free robot configuration exists for such a goal position.
In both cases, if no collision-free configurations are found, the cause of the failure is saved, increasing environment information and predicates for the task planner. 
A simple Knowledge Base module is in charge of memorising and updating such logical predicates, making them available to the planner.
In order to increase the efficiency and reduce computation time, valid robot configurations generated during the geometrical reasoning are stored in a cache and reused in subsequent geometrical reasoning executions, verifying that there are no collisions due to unexpected environmental changes.

\section{Experimental Results}
We evaluated our RH-TAMP approach in three classical planning problems: 
\begin{itemize}
    \item The \textit{Clutter Table} problem (e.g., \cite{srivastava2014combined,akbari2019combined})
    \item A variation of the \textit{Non-Monotonic} problem (e.g., \cite{lagriffoul2018platform})
    \item The \textit{Sort Clutter} problem (e.g., \cite{lagriffoul2018platform})
\end{itemize}
The actions used in these problems are the ones described in Section~\ref{sec:primitives}.

We used three robots simulated inside the MoveIt! environment, namely: 
\begin{itemize}
    \item A 7-DoF (Degrees of Freedom) Franka Emika Panda manipulator equipped with its standard, large parallel gripper
    \item A 6-DoF Universal Robots UR5 manipulator equipped with a Robotiq 2F-85 gripper
    \item A mobile manipulator composed of a Husky mobile base equipped with the UR5 manipulator described above.
\end{itemize}

In some experiments, we used both the manipulators listed above to show that our approach is effective regardless of the type of robot (7-DoF vs 6-DoF) and the size of the gripper.
All experiments were run on an Intel Core i7-770K 4.20 GHz CPU machine with 16 GB memory.
Some videos of the experiments can be found following this link:\\\\
\hspace{1cm}\url{https://nicolacastaman.github.io/rh-tamp/}

\subsection{Problems Definition}\label{sec:problems}
The selected benchmark problems satisfy some of the criteria defined in \cite{lagriffoul2018platform}:
\begin{itemize}
\item \textit{Infeasible task actions}: some actions are not feasible (i.e., no valid motion plan exists). Possible causes could be blocking objects and the kinematic limits of the robot.
\item \textit{Large task spaces}: the task planning problem requires a considerable search effort.
\item \textit{Non-monotonicity}: some objects may need to be moved more than once to reach the goal.
\end{itemize}

\subsubsection{Clutter Table}
\begin{figure}[!ht]
    \centering
    \begin{subfigure}{0.45\linewidth}
        \centering
        \includegraphics[width=\linewidth]{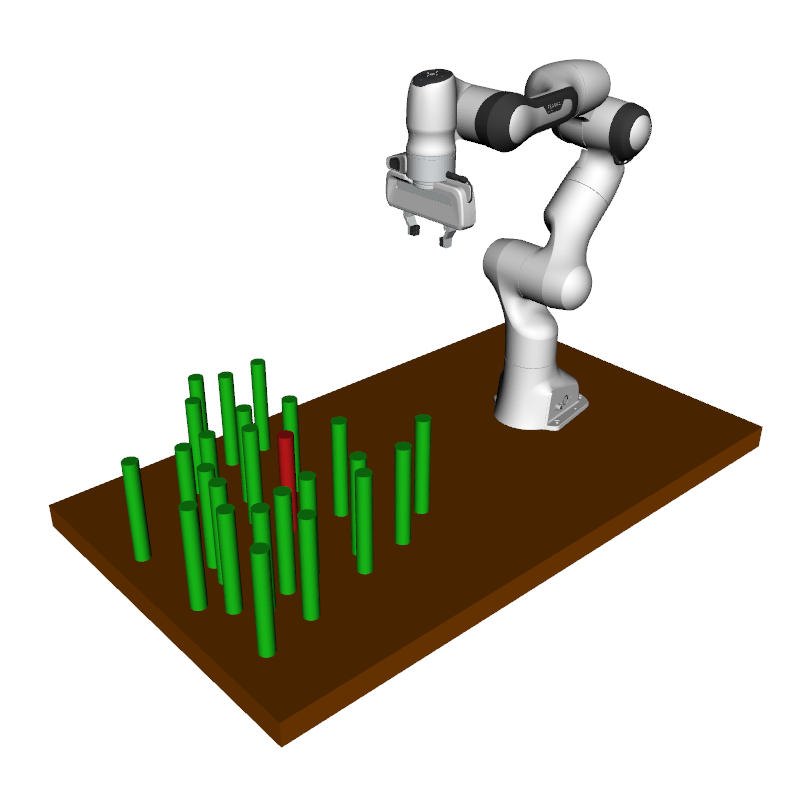}%
        \caption{Clutter Table problem with Franka Emika Panda.}
        \label{fig:problem1_panda}
    \end{subfigure}\hfill%
    \begin{subfigure}{0.45\linewidth}
        \centering
        \includegraphics[width=\linewidth]{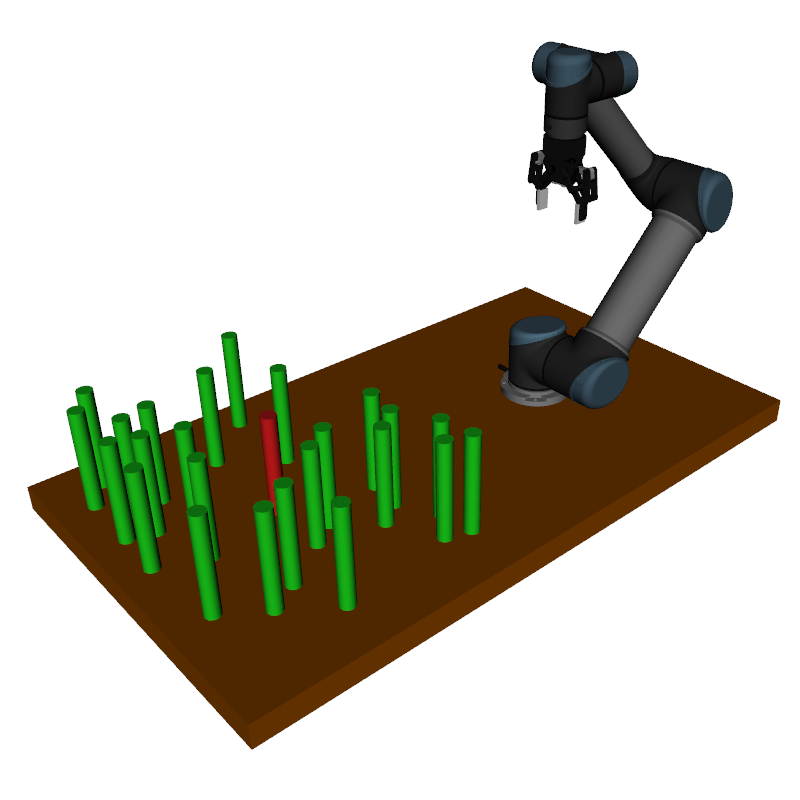}%
        \caption{Clutter Table problem with UR5.}
        \label{fig:problem1_ur}
    \end{subfigure}%
    \caption{Examples of the Clutter Table task and motion planning problem. The robot has to pick up the red cylinder placed between 25 other ones. The robot can grasp cylinders only from their side.}
    \label{fig:problem1}
\end{figure}

In the \textit{Clutter Table} problem, the robot has to pick up a specific cylinder (red cylinder in Fig.~\ref{fig:problem1}) from a table cluttered by many other cylinders (green cylinders in Fig.~\ref{fig:problem1}). The robot has to move
the other cylinders on the table to reach the target cylinder.    The robot is restricted to grasp cylinders only on their side. A \textit{Clutter Table} problem with 25 cylinders is depicted in Figure~\ref{fig:problem1}.
This problem evaluates the \textit{infeasible task actions} and the \textit{large task space criteria}. It requires a TAMP to carefully plan the sequence of actions and relocate green cylinders without creating new occlusions (infeasible task actions). In our experiments, different scenes are created, increasing the number of cylinders from 15 to 40, so solving the problem requires moving many objects, sometimes several times. This problem also evaluates the scalability of the proposed approach as the cardinality of the problem increases.

\subsubsection{Non-Monotonic}
\begin{figure}[!ht]
    \centering
    \includegraphics[width=0.45\linewidth]{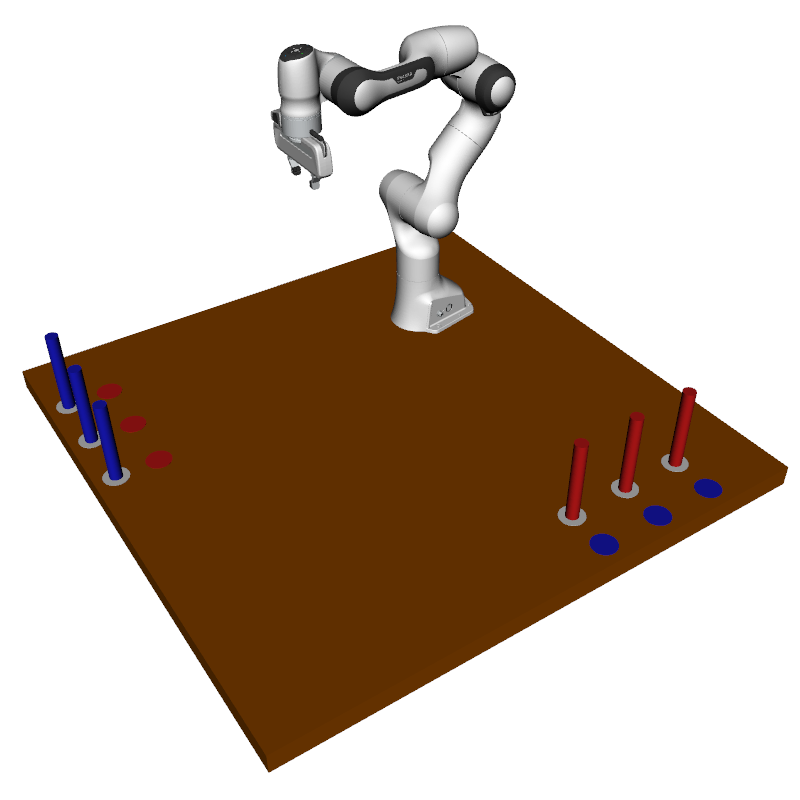}
    \caption{Example of a Non-Monotonic task and motion planning problem. The robot has to move the red and blue cylinders from their initial position to the positions corresponding to their colour. The robot can grasp cylinders only from their side.}
    \label{fig:problem2}
\end{figure}

In the variation of the \textit{Non-Monotonic} problem that we exploit in experiments, the robot has to move coloured cylinders (red and blue cylinders in Fig.~\ref{fig:problem2}) from their initial position to the positions corresponding to their colour. Our variation creates the condition in which a not optimized plan increases the number of actions necessary to complete the task (e.g. place a red cylinder in front of a blue one making it infeasible to pick up the blue cylinder). Also, in this case, the robot is restricted to grasp cylinders only from their side.
This problem evaluates the \textit{infeasible task actions} and \textit{non-monotonicity} criteria. 
For example, referring to Fig.~\ref{fig:problem2}, red cylinders block the blue ones' goal pose while red cylinders goal poses obstruct the pick up of the blue ones. Therefore, the goal condition of blue cylinders requires removing the red ones and bringing them to the final position after moving the blue ones to solve the problem.

\subsubsection{Sort Clutter}
\begin{figure}[!ht]
    \centering
    \includegraphics[width=0.7\linewidth]{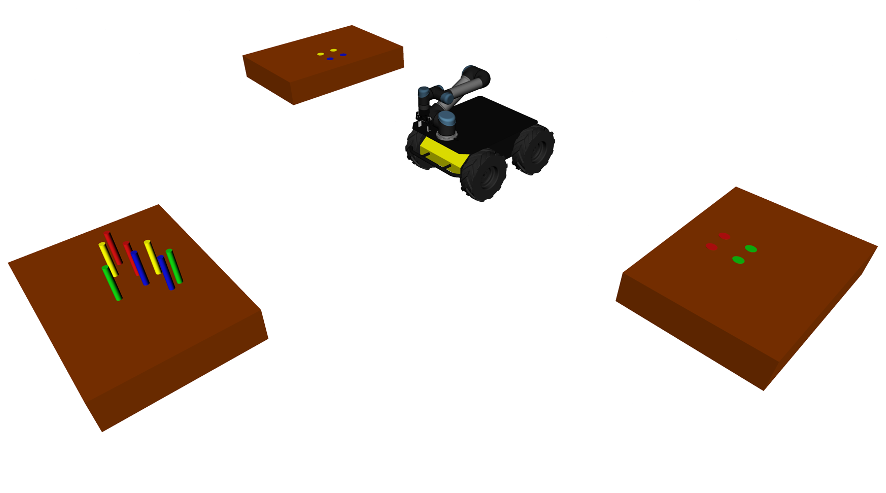}
    \caption{Example of a Sort Clutter task and motion planning problem. The mobile robot has to transport the cylinders of different colours from the initial position to the right table, sorting them by colour.}
    \label{fig:problem3}
\end{figure}

The third benchmark we present involves a particular instance of the \textit{Sort Clutter} problem. The robot has to move coloured cylinders, as in Fig.~\ref{fig:problem3}, from their initial position to the positions corresponding to their colour. To accomplish the task, the robot
must navigate and approach the tables.
Also, in this problem, a not optimized plan increases the number of actions necessary to complete the task (e.g., place a red cylinder before the green ones make it more difficult to place the latter).
The robot is equipped with a gripper and a tray, and it can grasp cylinders only from their side.
Consequently, some objects may block other objects, and some may block the target locations (\textit{infeasible task actions}). This will require temporarily move items away and bring them back later (\textit{non-monotonicity}).

\subsection{Metrics and Parameters}

As stated in~\cite{lagriffoul2018platform}, TAMP algorithms' performances can be measured in terms of planning time, success rate, or success rate within a time-bound. Also, the length of the computed plan both in terms of the number of actions and length of motion plan could be used.
In this work, we measure the performances of the proposed approach in terms of planning time, success rate within a time-bound.
All the experiments have been performed using the parameters reported in Table~\ref{tbl:parameters}.
\begin{table}[!ht]
\caption{Parameters used in all experiments}
\label{tbl:parameters}
\begin{center}
\begin{tabular}{ l | c}
  \toprule
    Parameters & Value\\
  \midrule 
    Max Task Planning Time & 10 sec\\
    Max Motion Planning Time & 10 sec\\
    Max replan in case of failure & 3\\
    Max planning time & 600 sec\\
    Max number of samples for geometric reasoning & 40\\
  \bottomrule
\end{tabular}
\end{center}
\end{table}

\subsection{Static Environment Results}
The proposed RH-TAMP approach is evaluated in a static environment on the \textit{Clutter Table} problem to compare with state-of-the-art TAMP algorithms. The obtained results are compared with two similar approaches that solve the same problem: Srivastava~\cite{srivastava2014combined} and Akbari~\cite{akbari2019combined}. We reproduced the same experiment performed in the mentioned papers, while the results of the compared approaches are taken from the respective papers. In this experiment, we used the \emph{Panda} robot.

\begin{table}[!ht]
\caption{Comparison of the success rate of our RH-TAMP with the approaches proposed in Srivastava~\cite{srivastava2014combined} and Akbari~\cite{akbari2019combined} on the \textit{Clutter Table} problem.}
\label{tbl:comparison_cluttered}
\begin{center}
\footnotesize
\begin{tabular}{ l|ccccc|c|c }
  \toprule
  \multirow{2}{*}{Problem} & \multicolumn{5}{c|}{Proposed} & \multirow{2}{*}{Srivastava~\cite{srivastava2014combined}} & \multirow{2}{*}{Akbari~\cite{akbari2019combined}} \\
  & h = 2 & h = 4 & h = 6 & h = 8 & h = inf & &\\
  \midrule 
  Clutter 15 & 100.0 & 100.0 & 100.0 & 100.0 & \textbf{100.0} & \textbf{100} & \textbf{100} \\
  Clutter 20 & 100.0 & 93.9  & 100.0 & 100.0 & \textbf{100.0} & 94  & \textbf{100} \\
  Clutter 25 & 100.0 & 100.0 & 100.0 & 100.0 & \textbf{100.0} & 90  & \textbf{100} \\
  Clutter 30 & 93.3  & 86.7  & 86.7  & 100.0 & 93.3  & 84  & \textbf{100} \\
  Clutter 35 & 100.0 & 93.9  & 100.0 & 100.0 & \textbf{100.0} & 67  & 95 \\
  Clutter 40 & 86.7  & 86.7  & 86.7  & 93.3  & 86.7  & 63  & \textbf{95} \\
  \bottomrule 
\end{tabular}
\end{center}
\end{table}

Table~\ref{tbl:comparison_cluttered} summarizes the results obtained by our RH-TAMP with different horizon values and compares them with approaches in~\cite{srivastava2014combined} and~\cite{akbari2019combined}; \emph{inf} means infinite action horizon, i.e., a horizon of the same size as the number of actions in the task plan.
The table shows that our approach outperforms the approach proposed in Srivastava while reaching performances comparable to Akbari.

\subsection{Changing Environment Results}

We tested RH-TAMP in a non-static environment on \textit{Clutter Table} and \textit{Non-Monotonic} problems.
To simulate the non-static nature and non-determinism of the environment, we generated unexpected object movements in the scene or grasping failures after a robot action, with a defined probability (defined below).
To compare our algorithm with classic TAMP approaches, the defined baseline corresponds to a state-of-the-art algorithm that plans the entire sequence of actions before executing them. In case of an unexpected event, it restarts the planning from the beginning considering the new state.

\subsubsection{Clutter Table}

RH-TAMP is first evaluated on the \textit{Clutter Table} problem. Then, the non-determinism of the environment is simulated, introducing an unexpected object movement or a grasping failure after each robot action with a probability of 20\%.

\begin{table}[!ht]
\caption{Comparison of the success rate with different Action Horizon on Clutter Table problem in Non-static Environments for two types of robots (Panda and UR5).}
\label{tbl:dynamic_rate_1}
\begin{center}
\footnotesize
\begin{tabular}{ c|l|c|ccccc }
  \toprule
  & \multirow{2}{*}{Problem} & \multicolumn{6}{c}{Success Rate (\%)}\\
  & & baseline & h = 2 & h = 4 & h = 6 & h = 8 & h = inf \\
  \midrule
  \multirow{6}{*}{\rotatebox[origin=c]{90}{Panda}} & Clutter 15 & 100.0 &  100.0 & 100.0 & 100.0 & 100.0 & 100.0 \\
  & Clutter 20 & 86.7 & 100.0 & 93.3 & 100.0 & 100.0 & 100.0 \\
  & Clutter 25 & 100.0 & 93.3 & 100.0 & 100.0 &  100.0 & 100.0 \\
  & Clutter 30 & 86.7 & 86.7 & 86.7 & 100.0 &  93.3 & 93.3 \\
  & Clutter 35 & 73.3 & 93.3 & 93.3 & 100.0 &  93.3 & 100.0 \\
  & Clutter 40 & 66.7 & 86.7 & 86.7 & 93.3 & 80.0 & 80.0 \\
  \midrule
  \multirow{6}{*}{\rotatebox[origin=c]{90}{UR5}} & Clutter 15 & 100.0 & 100.0 &  100.0 & 100.0 & 93.3 & 100.0 \\
  & Clutter 20 & 100.0 & 100.0 & 93.3 & 100.0 & 93.3 & 93.3 \\
  & Clutter 25 & 93.3 & 93.3 & 80.0 & 93.3 & 100.0 &  86.7 \\
  & Clutter 30 & 53.3 & 93.3 & 73.3 & 73.3 & 80.0 &  80.0 \\
  & Clutter 35 & 73.3 & 73.3 & 80.0 & 73.3 & 73.3 &  86.7 \\
  & Clutter 40 & 53.3 & 66.7 & 80.0 & 66.7 & 66.7 & 66.7 \\
  \bottomrule
\end{tabular}
\end{center}
\end{table}

Table~\ref{tbl:dynamic_rate_1} reports the obtained success rate with different task horizon size compared with the baseline. Results demonstrate that the RH-TAMP approach outperforms the baseline independently from the selected horizon in most of the cases. Moreover, results suggest that an action horizon between 6 and 8 guarantees a slightly higher success rate. This is an expected behaviour. Indeed higher action horizon allows optimizing of the sequence of actions that has to be performed.

\begin{table}[!ht]
\caption{Comparison of the average execution time (panning time + execution time) with different Action Horizon on Clutter Table problem in Non-static Environments for two types of robots (Panda and UR5). The time corresponds to the time robot takes to complete the task.}
\label{tbl:dynamic_time_1}
\begin{center}
\footnotesize
\begin{tabular}{ c|l|c|ccccc }
  \toprule
  & \multirow{2}{*}{Problem} & \multicolumn{6}{c}{Avg Time (s)}\\
  & & baseline & h = 2 & h = 4 & h = 6 & h = 8 & h = inf \\
  \midrule
  \multirow{6}{*}{\rotatebox[origin=c]{90}{Panda}} & Clutter 15 & 29.190 & 27.007 & 21.198 & 25.903 & 22.004 & 22.271\\
  & Clutter 20 & 47.512 & 59.616 & 35.737 & 48.605 & 57.413 & 67.869\\
  & Clutter 25 & 104.083 & 53.199 & 61.770 & 65.884 & 65.035 & 54.595\\
  & Clutter 30 & 187.603 & 86.077 & 74.340 & 98.477 & 109.446 & 93.694\\
  & Clutter 35 & 174.468 & 104.25 & 113.864 & 118.759 & 106.435 & 109.092\\
  & Clutter 40 & 276.404 & 177.801 & 180.925 & 200.975 & 154.914 & 156.065\\
  \midrule
  \multirow{6}{*}{\rotatebox[origin=c]{90}{UR5}} & Clutter 15 & 46.079 & 34.886 & 26.546 & 27.768 & 34.877 & 28.680\\
  & Clutter 20 &  159.615 & 72.459 & 61.804 & 69.311 & 53.024 & 54.810 \\
  & Clutter 25 &  149.179 & 74.153 & 62.608 & 85.358 & 75.176 & 72.284 \\
  & Clutter 30 &  374.763 & 159.354 & 135.334 & 166.528 & 134.598 & 168.655\\
  & Clutter 35 &  344.123 & 121.393 & 140.312 & 174.337 & 193.314 & 145.660 \\
  & Clutter 40 &  306.647 & 209.676 & 199.910 & 266.231 & 184.038 & 287.590 \\
  \bottomrule
\end{tabular}
\end{center}
\end{table}

\begin{figure}[!ht]
    \centering
    \begin{subfigure}{0.49\linewidth}
        \centering
        \includegraphics[width=\linewidth]{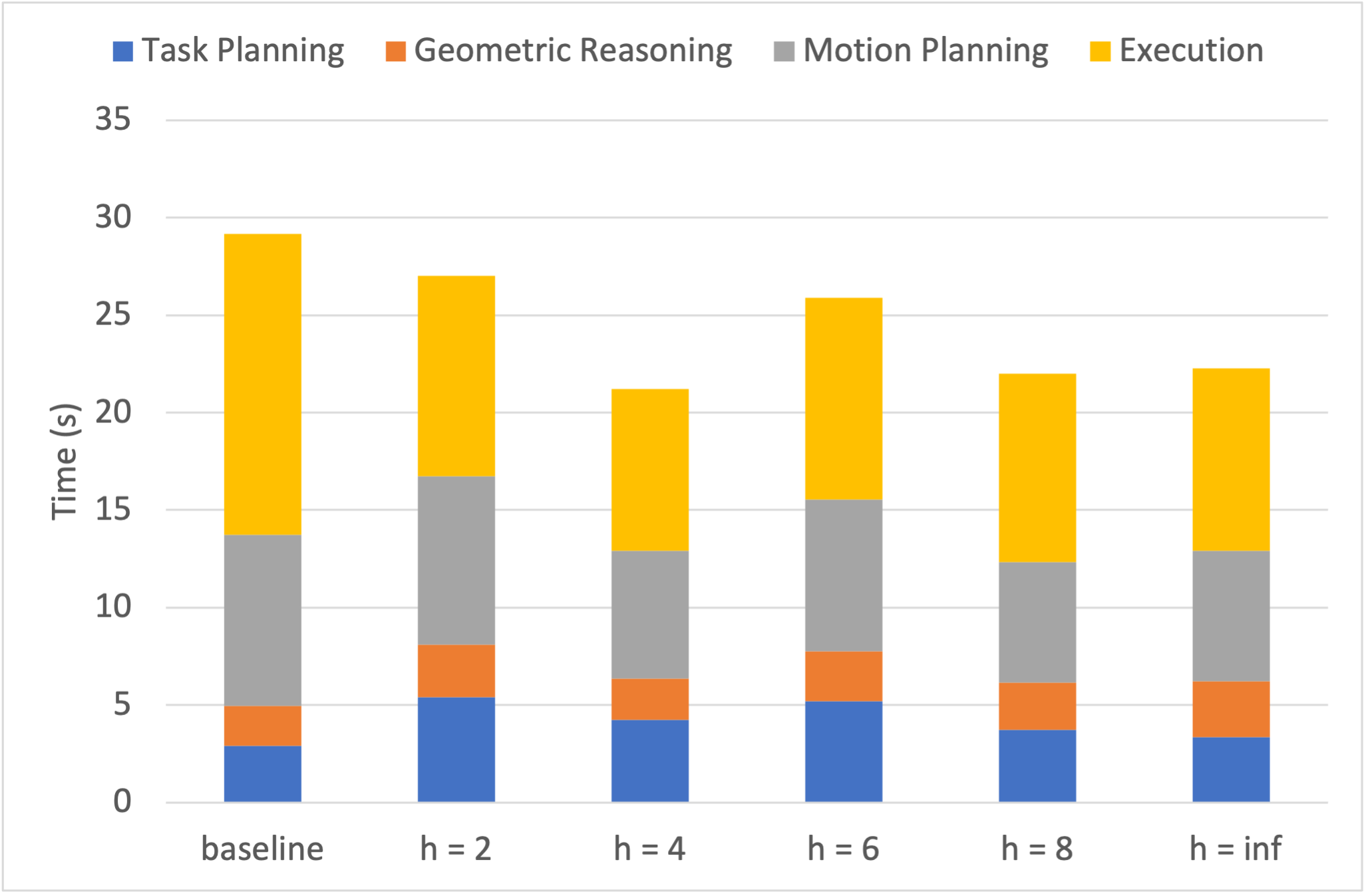}%
        \caption{Clutter 15}
        \label{fig:1}
    \end{subfigure}\hfill%
    \begin{subfigure}{0.49\linewidth}
        \centering
        \includegraphics[width=\linewidth]{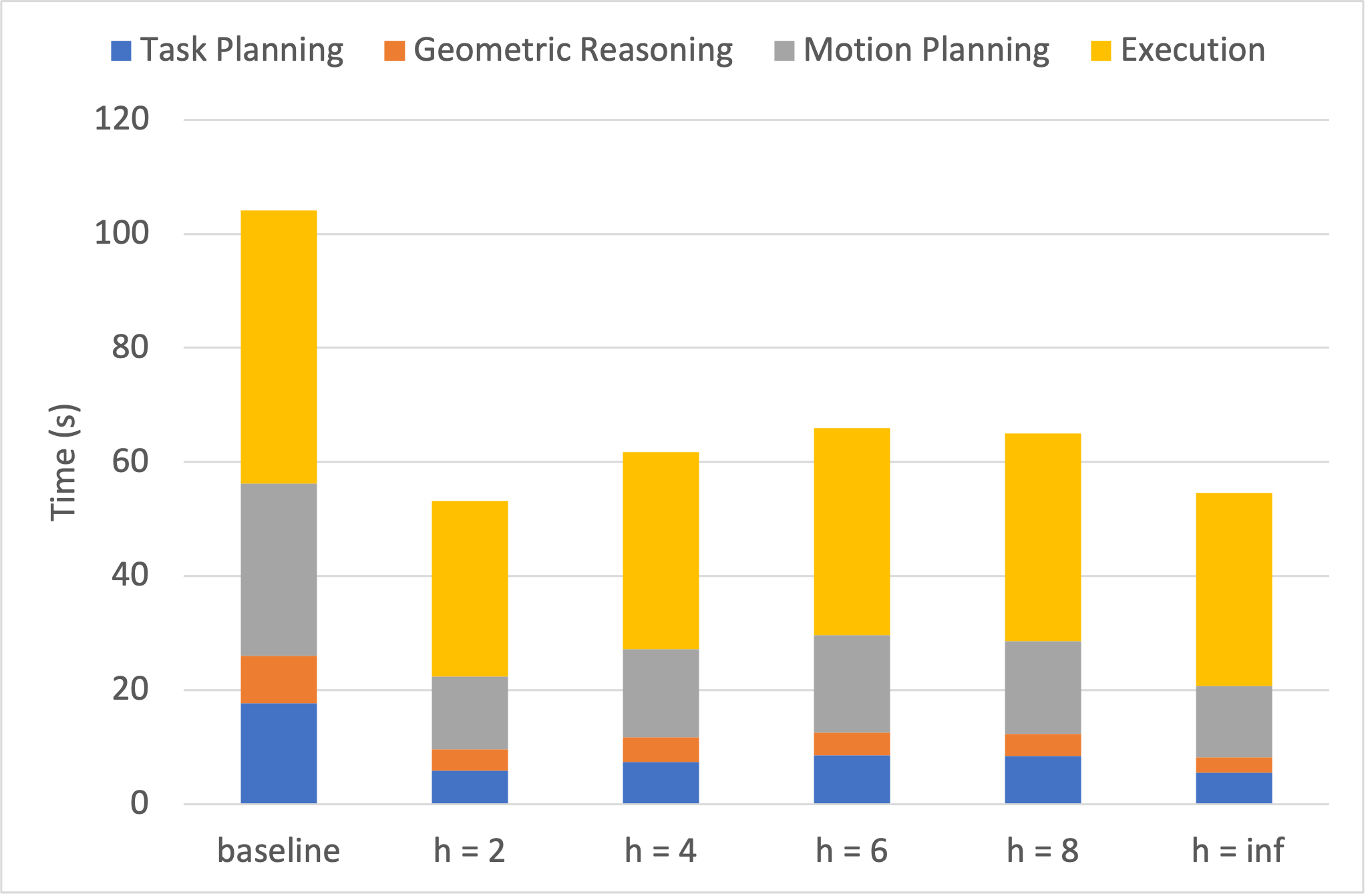}%
        \caption{Clutter 25}
        \label{fig:2}
    \end{subfigure}
    \par\bigskip
    \begin{subfigure}{0.49\linewidth}
        \centering
        \includegraphics[width=\linewidth]{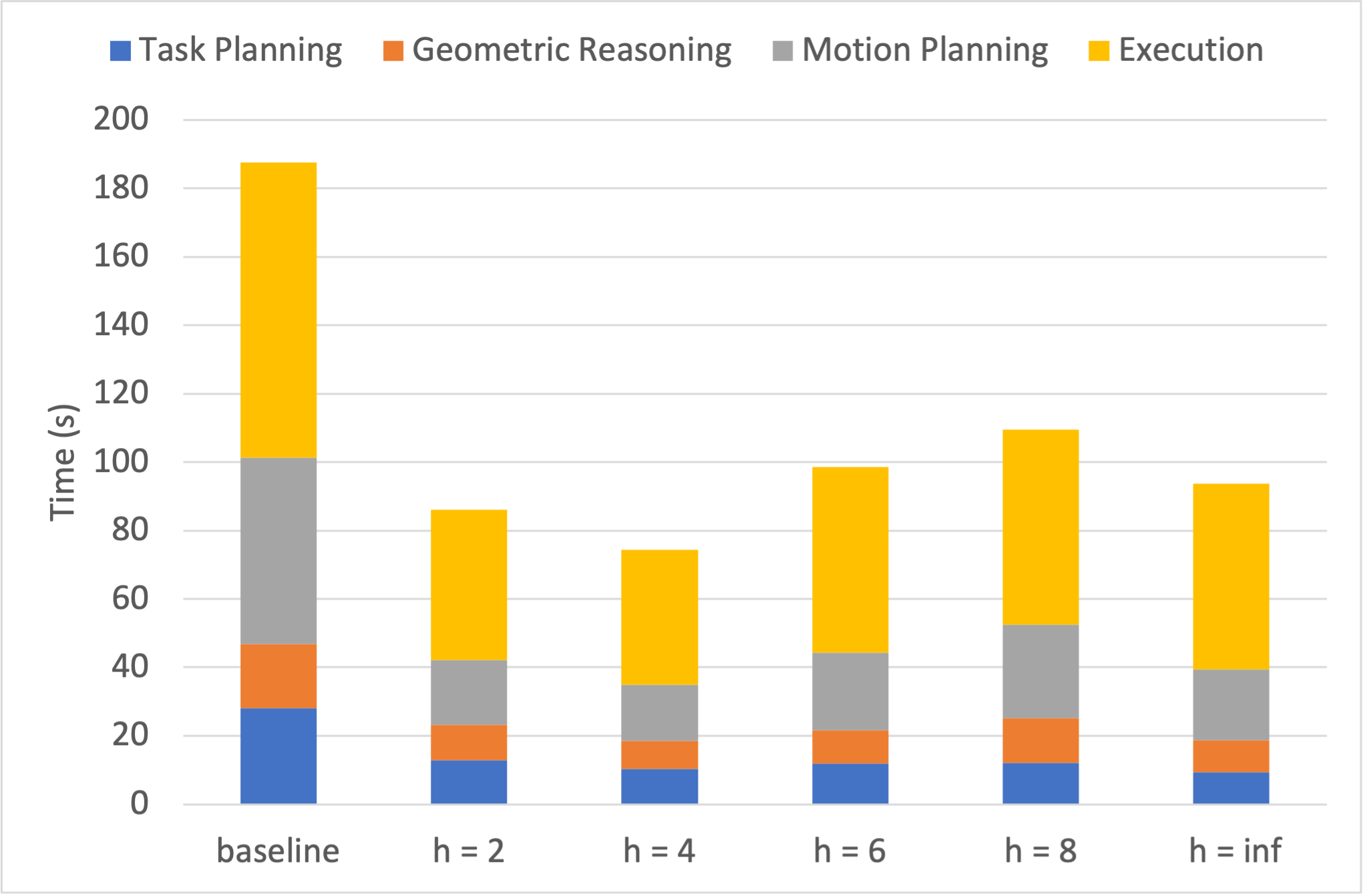}%
        \caption{Clutter 30}
        \label{fig:3}
    \end{subfigure}\hfill%
    \begin{subfigure}{0.49\linewidth}
        \centering
        \includegraphics[width=\linewidth]{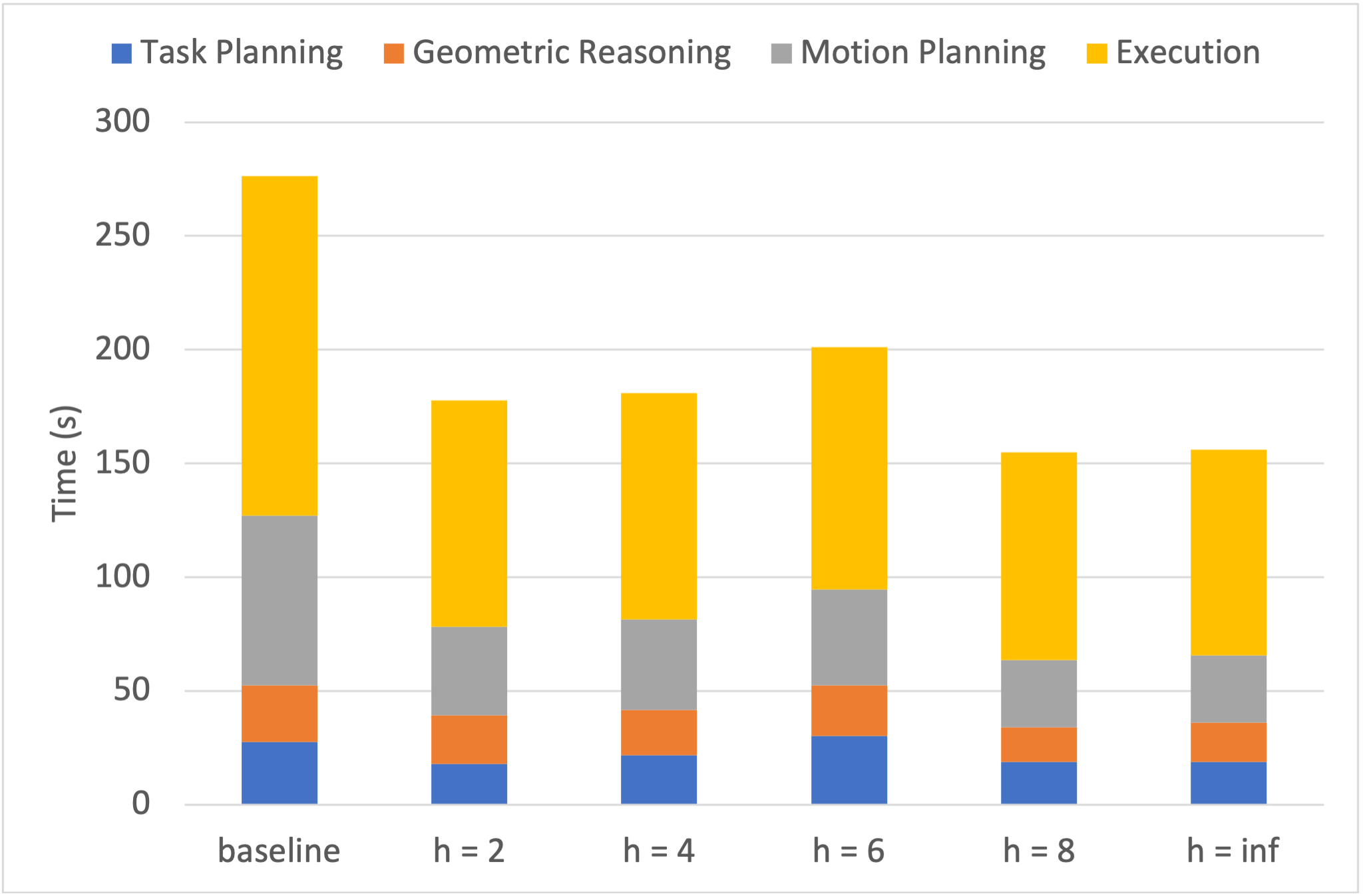}%
        \caption{Clutter 40}
        \label{fig:4}
    \end{subfigure}%
    \caption{Average time spent by each component of the pipeline: task planning, geometric reasoning, motion planning, and execution for some of the benchmarks. For each benchmark, the bins of the histograms refer to a different horizon.}
    \label{fig:graphs}
\end{figure}

Table~\ref{tbl:dynamic_time_1} reports the average execution time compared with the baseline. Presented time values correspond to the time necessary to complete the task (i.e. planning and execution). It is not possible to separate planning and execution due to the behaviour of the proposed algorithm.
Obtained results demonstrate that RH-TAMP has an execution time of around 50\% less time than the baseline.

Figure~\ref{fig:graphs} reports the average time spent running each module, namely task planning, geometric reasoning, motion planning, and action execution, for some of the presented experiments. From the plots, we can deduce that more than 50\% of the time is dedicated to the execution of the trajectories. With respect to the planning time, the motion planner is the most expensive task that takes up 20\% of the time. However, thanks to the receding horizon approach, the time spent on motion planning is clearly less than in the baseline. Moreover, the geometric reasoning avoids calling on motion planning of invalid configuration. Indeed, it finds invalid configurations about 50\% of times while spending only 10\% of the total time.

\subsubsection{Non-Monotonic}

The \textit{Non-Monotonic} problem tests the planner producing situations where bad planning increases the number of actions to achieve the goal (e.g., placing a red cylinder on its goal position without removing it before the corresponding blue cylinder). In this problem, the non-determinism of the environment is simulated, introducing an unexpected object movement after each robot action with a probability of 10\%. 
This problem aims to highlight the advantage of using a slightly longer horizon.

\begin{table}[ht]
\caption{Comparison of the average execution time (panning time + execution time) with different Action Horizon on \textit{Non-monotonic} problem in Non-static Environments for two types of robots (Panda and UR5). The time corresponds to the time robot takes to complete the task.}
\label{tbl:dynamic_2}
\begin{center}
\footnotesize
\begin{tabular}{ c|l|c|ccccc }
  \toprule
  & \multirow{2}{*}{Problem} & \multicolumn{6}{c}{Avg Time (s)}\\
  & & baseline & h = 2 & h = 4 & h = 6 & h = 8 & h = inf \\
  \midrule
  \rotatebox[origin=c]{90}{Panda} & Non-Monotonic & 177.472 & 82.412 & 79.063 & 70.890 & \textbf{69.608} & 75.402\\
  \midrule
  \rotatebox[origin=c]{90}{UR5} & Non-Monotonic & 256.987 & 144.111 & 142.960 & \textbf{138.133} & 140.344 & 147.358\\
  \bottomrule
\end{tabular}
\end{center}
\end{table}

The proposed RH-TAMP reached a 100\% success rate in all execution.
Table~\ref{tbl:dynamic_2} presents the average execution time and the comparison with the baseline values. Also, the presented time values correspond to the time necessary to complete the task (i.e. planning and execution) for this problem.
Results also confirm in this case that RH-TAMP has an execution time of around 50\% less than the baseline.

We observe that the best results are obtained on average with a horizon of 6 or 8 with both the robots.
Indeed, a slightly wider horizon helps optimize the solution, which is critical because the goal position of red cylinders obstructs the blue ones. 
Planning over a long actions horizon allows optimizing the sequence of actions, thus reducing the number of total actions and consequently the planning and execution times.
On the other hand, it is important to remark that the action horizon choice is influenced by the probability of changes in the environment. Indeed, an environment that changes configuration very often makes the optimization obtained using a large action horizon useless. In such conditions, a different trade-off between the length of the horizon and the responsiveness of the system should be found.

\begin{figure}[ht]
    \centering
    \includegraphics[width=0.9\linewidth]{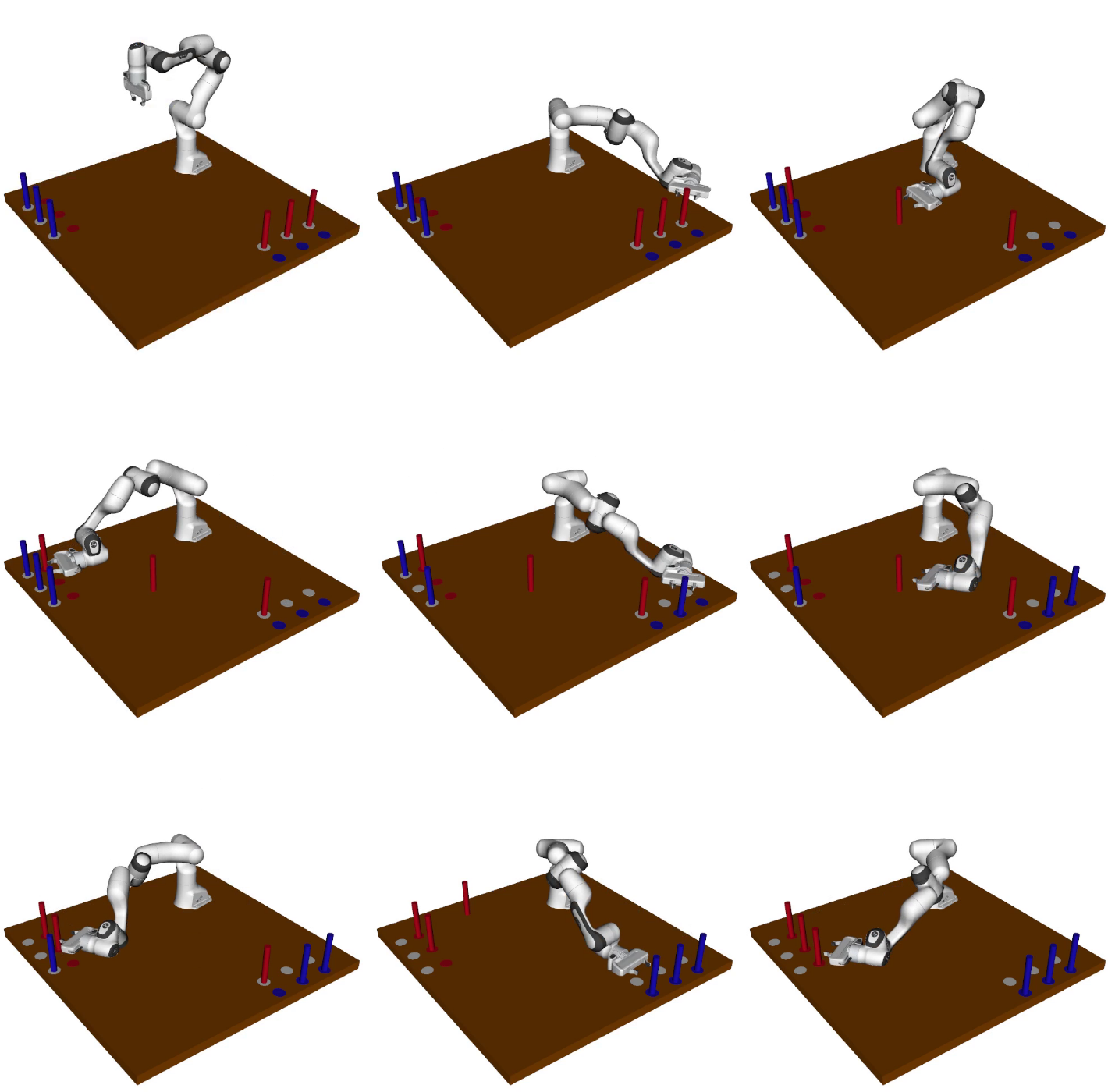}
    \caption{An example sequence of the execution of the Non-Monotonic problem.}
    \label{fig:problem2_sequence}
\end{figure}

An example sequence of the execution of the problem is depicted in Fig.~\ref{fig:problem2_sequence} while a video could be found at \url{https://nicolacastaman.github.io/rh-tamp/}.

\subsubsection{Sort Clutter}
The \textit{Sort Clutter} problem also tests  our system'se adaptability to localization uncertainties (e.g., the robot may not accurately dock near the tables).
The performance is measured in terms of average planning time and success rate within a time-bound. 
For each experiment, we performed 15 trials solving the \textit{Sort Clutter} problem with the parameters of Table~\ref{tbl:parameters}.
To simulate the non-static nature of the environment, we generated an unexpected object's movement after each robot action with a 20\% probability. To compare our proposal with classic state-of-the-art TAMP algorithms, we defined a \textit{baseline} approach that plans the entire action sequence before its execution. In case of unexpected events, it restarts the planning from the beginning while considering the new state.

\begin{table*}[ht]
\caption{Comparison of the average execution time (panning time + execution time) with different Action Horizon on \textit{Sort Clutter} problem in Non-static Environments. The time corresponds to the time robot takes to complete the task.}
\label{tbl:results}
\begin{center}
\footnotesize
\begin{tabular}{ l|c|ccccc }
  \toprule
   Problem & baseline & h = 2 & h = 4 & h = 6 & h = 8 & h = inf \\
  \midrule
   4 Cylinders & 315.275 & 173.738 & 166.241 & \textbf{154.194} & 161.905 & 171.113\\
   8 Cylinders & 715.598 & 386.109 & 399.549 & \textbf{357.371} & 363.137 & 372.583\\
   12 Cylinders & 1129.959 & 817.346 & 800.315 & 804.701 & \textbf{787.913} & 798.108\\
  \bottomrule
\end{tabular}
\end{center}
\end{table*}

Table~\ref{tbl:results} shows obtained results for $M=\{4, 8, 12\}$ cylinders of different colours. Time values include both the planning and execution time. As shown, for $M=4$, \textit{baseline} takes 315.275s to complete the assignment, while \textit{RH-TAMP} with $h=4$ takes 154.194s. 
For $M=12$, 787.913s of \textit{RH-TAMP} ($h=8$) are compared with the 1129.959s of \textit{baseline}. On average, the proposed approach is twice as faster as \textit{the baseline}.

\begin{table}[t]
\caption{Comparison between success rates of \textit{baseline} and \textit{RH-TAMP} with different horizons.}
\label{tbl:results_2}
\begin{center}
\footnotesize
\begin{tabular}{ l|c|ccccc }
  \toprule
   Problem & baseline & h = 2 & h = 4 & h = 6 & h = 8 & h = inf \\
  \midrule
   4 Cylinders & 100.0 & 100.0 & 100.0 & 100.0 & 100.0 & 100.0\\
   8 Cylinders & 93.3 & 100.0 & 93.3 & 100.0 & 100.0 & 100.0\\
   12 Cylinders & 66.7 & 86.7 & 93.3 & 100.0 & 93.3 & 100.0\\
  \bottomrule
\end{tabular}
\end{center}
\end{table}

Table~\ref{tbl:results_2} shows the success rate. Again, different horizons are reported ($h=\{2, 4, 6, 8, inf\}$), together with a different number of objects to be sorted ($M=\{4, 8, 12\}$). Results demonstrate that \textit{RH-TAMP} outperforms the baseline independently from the selected horizon in most of the cases. Similarly to the manipulation setup, results suggest that an action horizon between 6 and 8 guarantees a slightly higher success rate. Overall, the reported results show that the proposed approach is effective regardless of the robot's scenario and the number of its degrees of freedom.

\begin{figure}[ht]
    \centering
    \includegraphics[width=\linewidth]{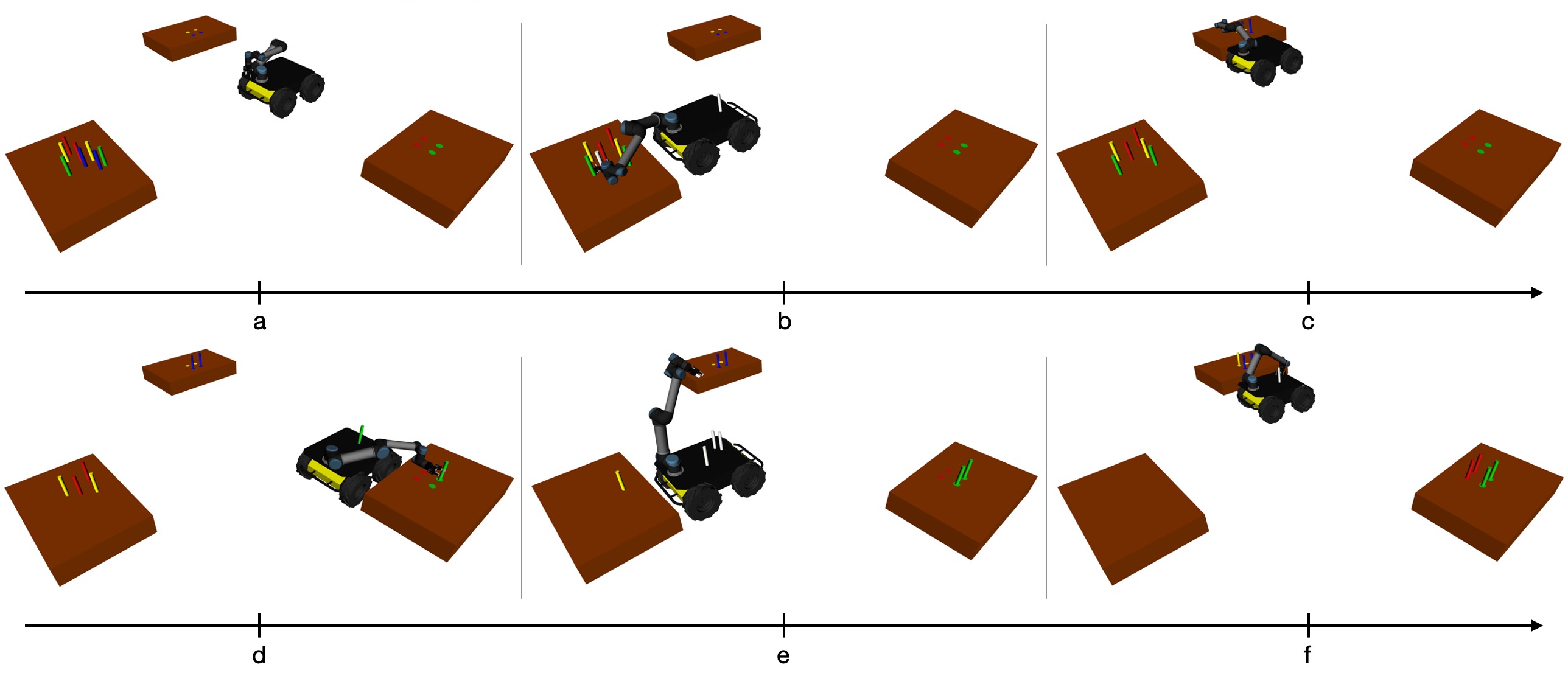}
    \caption{An example sequence of the execution of the Sort Clutter problem.}
    \label{fig:problem3_sequence}
\end{figure}

Figure \ref{fig:problem3_sequence} shows an example of the simulated execution sequence for 8 cylinders of 4 colours.

\section{Conclusions}
In this paper, we present a novel approach called \emph{Receding Horizon Task and Motion Planning} (RH-TAMP) to solve Task and Motion Planning problems taking into account non-deterministic actions and changing environments. The key idea of our approach is to efficiently solve at each iteration a reduced TAMP problem over a receding horizon of future actions, scheduling only the first action of the horizon. We validated our method within extensive simulated experiments on three different TAMP benchmarks with three different robots. Our results suggest that our RH-TAMP is able to handle non-static TAMP problems regardless of their cardinality while ensuring comparable performance with respect to other recent TAMP approaches in solving traditional, static problems. Furthermore, we showed that in a changing environment, our approach in most cases allows us to solve TAMP problems with a higher success rate and in less time than a standard approach, reducing the runtime by up to 50\%.
The open-source implementation of our system is made publicly available with this paper.\\

As future work, we plan to test the proposed approach with a real vision-guided robot and to deploy a more generalized and exploitable version of our TAMP simulation framework.

\section*{Acknowledgements}
MIUR (Italian Minister for Education) supported part of this work under the initiative ``Departments of Excellence'' (Law 232/2016).

\bibliographystyle{elsarticle-num} 
\bibliography{task,motion,tamp}

\end{document}